\documentclass[letterpaper, 10 pt, conference]{ieeeconf}  % Comment this line out if you need a4paper

\IEEEoverridecommandlockouts                              % This command is only needed if 
\usepackage{graphicx} % for pdf, bitmapped graphics files
\usepackage{multirow}
\usepackage{nicefrac, xfrac}
\usepackage{amsmath} % assumes amsmath package installed
\usepackage{amssymb}  % assumes amsmath package installed
\usepackage{algpseudocode}
\usepackage{xcolor}
\usepackage{lipsum}
\usepackage{booktabs}
\usepackage{adjustbox}
\usepackage{mathtools}
\usepackage{float}
\usepackage{amssymb}% http://ctan.org/pkg/amssymb
\usepackage{pifont}% http://ctan.org/pkg/pifont
\newcommand{\cmark}{\ding{51}}%
\newcommand{\xmark}{\ding{55}}%

\let\labelindent\relax
\usepackage{enumitem}

\setlist[itemize]{leftmargin=10pt}
\definecolor{ForestGreen}{rgb}{0.133, 0.545, 0.133}
\usepackage[ruled,vlined,linesnumbered]{algorithm2e}

\SetCommentSty{mycommfont}
\title{\LARGE \bf
Real-to-Sim Deformable Object Manipulation: Optimizing Physics Models with Residual Mappings for Robotic Surgery
}

\author{
Xiao Liang$^\dagger$, Fei Liu$^\dagger$, Yutong Zhang, Yuelei Li, Shan Lin, and Michael Yip \IEEEmembership{Senior Member, IEEE}% <-this % stops a space
\thanks{$^\dagger$ Equal contributions.}% <-this % stops a space
\thanks{Authors are affiliated with Department of Electrical and Computer Engineering, University of California San Diego, La Jolla, CA 92093 USA.{\tt\small\{x5liang, f4liu, yuz049, yul189, shl102, yip\}@ucsd.edu}}%
}

\begin{document}

\maketitle
\thispagestyle{empty}
\pagestyle{empty}

%%%%%%%%%%%%%%%%%%%%%%%%%%%%%%%%%%%%%%%%%%%%%%%%%%%%%%%%%%%%%%%%%%%%%%%%%%%%%%%%
\begin{abstract}
Accurate deformable object manipulation (DOM) is essential for achieving autonomy in robotic surgery, where soft tissues are being displaced, stretched, and dissected. Many DOM methods can be powered by simulation, which ensures realistic deformation by adhering to the governing physical constraints and allowing for model prediction and control. However, real soft objects in robotic surgery, such as membranes and soft tissues, have complex, anisotropic physical parameters that a simulation with simple initialization from cameras may not fully capture. To use the simulation techniques in real surgical tasks, the real-to-sim gap needs to be properly compensated. In this work, we propose an online, adaptive parameter tuning approach for simulation optimization that (1) bridges the real-to-sim gap between a physics simulation and observations obtained 3D perceptions through estimating a residual mapping and (2) optimizes its stiffness parameters online. Our method ensures a small residual gap between the simulation and observation and improves the simulation's predictive capabilities. The effectiveness of the proposed mechanism is evaluated in the manipulation of both a thin-shell and volumetric tissue, representative of most tissue scenarios. This work contributes to the advancement of simulation-based deformable tissue manipulation and holds potential for improving surgical autonomy.

% two real-world tasks: (a) manipulation of a thin-shell soft object, 2. manipulation of a piece of tissue. 

% Deformable simulation, that guarantees realistic interaction and deformation by respecting the governing physics, can potentially power a wide range of robot DOM algorithms. 

\end{abstract}

%%%%%%%%%%%%%%%%%%%%%%%%%%%%%%%%%%%%%%%%%%%%%%%%%%%%%%%%%%%%%%%%%%%%%%%%%%%%%%%%

\section{INTRODUCTION}
Handling deformable objects is a fundamental skill in robotic surgery, where surgeons precisely and minimally invasively treat soft tissue disease with robots. As the scope of robotic surgery expands, and as the gap between trained doctors and under-served patient populations widen, AI-driven surgical capabilities like autonomous suturing and tissue retraction become increasingly desirable and potentially necessary~\cite{Yip_2023, Attanasio_2021}. Ultimately, these robots will require an understanding of tissue physics and deformable object manipulation in general in order to provide assistance or take over a surgical task.
 
Physics-based simulations have been proven to be a promising technique for deformable object manipulation (DOM) \cite{Arriola_2020, Jihong_Zhu_2022, Jinao_2018, Collins_2021}. Extensive research has been done on modeling, motion planning, and data representations for perception perspectives \cite{Hang_Yin_2021}. The development of physical simulation environments can support a variety of deformable objects, including thin-shell fabric \cite{Qingyang_2020_RAL}, linear elastic ropes \cite{Fei_2023_RAL}, volumetric tissue \cite{Fei_2021_ICRA, Yunhai_2020}, and fluids \cite{Jingbin_2021_ICRA}.

Considering many existing deformable simulators, the high computational cost and ``reality gap" are two factors that limit the conventional simulation approaches. These include Finite Element Methods (FEM) and mass-spring system \cite{Faure_2012_SOFA}. Uncertainties, inaccurately calibrated parameters, and unmodeled physical effects can all lead to the gap between simulation and reality. As a result, many simulators need their parameters fine-tuned \textit{online} in order to be deployable for real-world robotic tasks. With the intention of bridging the gap between reality and simulation, the phrase real-to-sim was first used in \cite{Fei_2021_ICRA} to describe approaches that compensate for the simulated errors from live observations.
\begin{figure}[t]
    \centering
    \includegraphics[width=0.9\linewidth]{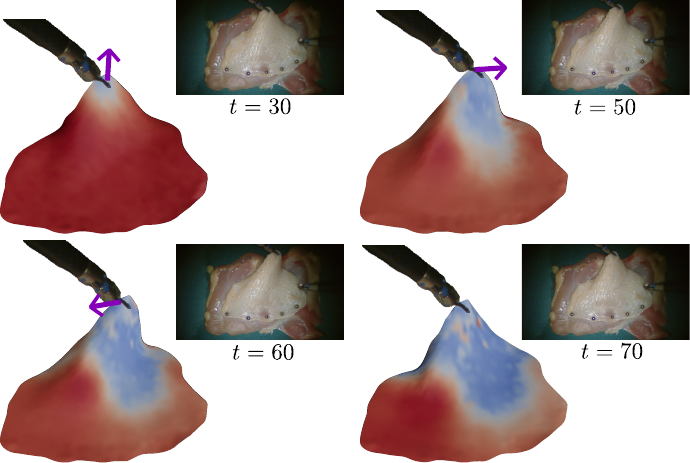} 
    \caption{Our proposed online simulation parameter optimization method. It reveals suitable constraints' stiffness parameters for a PBD simulation of a deformable tissue in an online manner. Arrows in purple shows the immediate control direction at the given time step. The tissue's blue regions indicate lower stiffness, whereas red regions have higher stiffness. In this figure, the tissue is initialized with large stiffness parameters.}
    \label{fig:coverphoto}
    \vspace{-1.5em}
\end{figure}

Our goal is to investigate the real-to-sim gaps by taking into account both known physical (geometric, mechanical) constraints describing tissue deformation and observable high-dimensional data, i.e., point cloud. For real-time applications such as surgical tissue manipulation, grasping, and retraction, it would also be necessary to identify the proper simulation parameters of soft bodies with a fast online simulation approach.

\subsection{Related Works}
The real-to-sim problem has recently gained attention in the literature. Most of the existing works have been focusing on closing the gap through effective policy transfer in a reinforcement learning (RL) setup, as reviewed in \cite{Wenshuai_2020, Salvato_2021}. Most of these works only rely on simulating rigid objects in the scene and robots with kinematics \cite{Chebotar_2019, 9706346}. Recently, several papers used similar deep RL for deformable objects, such as cloth \cite{Matas_2018}, tissue \cite{Scheikl_2023}, and ropes \cite{Yuqing_2021_ICRA}. However, these works don't create explicit physical models but learn the system parameters or controls in an end-to-end manner. It leads to several typical problems with learning strategies, including generalizability and data-hunger. 

Other methods rely on physics-based simulation to minimize the real-to-sim gap. Table~\ref{tbl:previsou_work_table} presents a summary. In our previous paper \cite{Fei_2021_ICRA}, we directly update the simulated positions of the volumetric particles using the spatial gradient of the signed distance field of point cloud observation. However, the simulation parameters are not updated, necessitating frame-by-frame registration. Similarly, \cite{sundaresan2022diffcloud} optimizes for a simulation parameter, such as mass or stiffness, to minimize the difference between a simulation and the point cloud observation. Furthermore, the authors enhanced their approach in \cite{Antonova_2022_RAL} as probabilistic inference over simulation parameters of the deformable object. However, their method is limited to thin-shell or linear objects (cloth, rope) with surface point clouds. Projective dynamics \cite{Afshar_2022_TMECH} were used to create a real-time physics-based model for tissue deformation, enhanced by a Kalman filter (KF) for refining the simulation with surface marker data. In a related work \cite{Yafei_2023_RAL}, they integrated FEM-based simulation into a deep reinforcement learning framework for grasping point policy learning, relying on offline stereo calibration for registration with the real world. For further improvements, \cite{Afshar_2023_TMECH} utilized a variational autoencoder with graph-neural networks to learn low-dimensional latent state variables' probability distributions. These variables were iteratively updated using an ensemble smoother with data assimilation to align the simulation with real data. However, their offline training for specific FEM simulation datasets poses challenges in real-world applications, especially in surgery, where lengthy data-collection and pretraining phases are impractical.

% Furthermore, projective dynamics \cite{Afshar_2022_TMECH} were employed to create a real-time physics-based model for tissue deformation, complemented by a Kalman filter (KF) for predicting and refining the simulation model with the captured marker data on the surface tissue. In \cite{Yafei_2023_RAL}, the FEM-based simulation is integrated into a deep reinforcement learning framework for optimal policy learning of a grasping point. To register the simulation gap with the real world, they only used offline stereo calibration before experimentation without online updating. For enhancements, \cite{Afshar_2023_TMECH} uses a variational autoencoder with graph-neural networks to learn the probability distributions of low-dimensional latent state variables. Then, it is updated iteratively using an ensemble smoother with multiple data assimilation in order to match the simulation with real data. However, their low-dimensional latent space representation requires offline training for specific datasets of FEM simulation. These offline techniques are difficult to realize in the real world and especially in surgery, given that task execution in real life rarely affords lengthy data-collection pretraining phases on-site. 
% It may difficult to preserve the original physical properties regardless of the registration errors. 

\begin{table}[t!]
\setlength\tabcolsep{1em}
\centering
\caption{summary of selected previous works that addressed deformable real-to-sim gap. }
\begin{adjustbox}{width=0.45\textwidth}
    \begin{tabular}{l|ccc}
    \toprule
    % \multirow{2}{*}{Velocity}
 
 Methods & Residual & Parameters Update  & Object types\\ 
  \midrule
  \cite{Fei_2021_ICRA} & \cmark & \xmark& volumetric\\
\cite{sundaresan2022diffcloud}& \xmark& \cmark  & thin-shell\\
\cite{Afshar_2023_TMECH}&\cmark & \xmark  & volumetric\\
Proposed & \cmark & \cmark & both \\
    \bottomrule
\end{tabular}
\end{adjustbox}
\label{tbl:previsou_work_table}
\vspace{-1.5em}
\end{table}
% \begin{itemize}
%     \item \cite{sundaresan2022diffcloud} optimizes for simulation parameter to minimize the difference between a simulation and point cloud observation. The method requires online optimization and works only with simple real-world object.
%     \item \cite{Yafei_2023_RAL} presents a sim-to-real framework for using deep reinforcement learning (DRL) in the FEM-based simulation. However, their simulated environment is simple compared to other complex real robotic setups. To register the simulation with the real world, they only used offline stereo calibration before experiments.
%     \item \cite{Afshar_2023_TMECH} The proposed method use a variational autoencoder (VAE) with graph-neural networks to learn the probability distributions of low-dimensional latent variables. Then it is updated iteratively using an ensemble smoother with multiple data assimilation (ES-MDA), in order to match simulation with real data. However, their low-dimensional latent space representation requires off-line training for specific datasets of FEM simulation. It may difficult to preserve the original physical properties regardless of the registration errors.
% \end{itemize}
\subsection{Contributions}
% As a summary, none of previous works has the ability to infer morphology of a real deformable object, as well considering the physical constraints.
In this work, we propose an online simulation parameter optimization framework for bridging the real-to-sim gap in deformable object manipulation. We believe it offers significant benefits in recovering a realistic physics simulation model for planning and control in DOM scenarios, including surgical autonomy. To this end, we present the following novel contributions:
\begin{itemize}
    \item An online ``real-to-sim'' residual mapping module is seamlessly incorporated into a physics simulation loop to predict a residual deformation $\Delta_t$ over time, which modifies simulation states to align with point cloud observations.
    \item The module estimates geometry-aware deformation for sub-surface simulation particles by considering geometric constraints. Therefore, it can be applied to thin-shell and volumetric deformable objects, covering wider surgical use cases than previous works.
    \item An online simulation optimization framework is used that adaptively updates stiffness parameters through loss functions that are informed by the residual mapping module, enhancing the simulation's ability to predict future deformation and minimizing the the ``real-to-sim'' gap.
    % \item A 3D reconstruction pipeline is developed to estimate point cloud observations of tissues. We evaluate the proposed framework in multiple real tissue manipulation scenarios and prove its effectiveness.
\end{itemize}

% conduct real-world deformable manipulation experiments on two different types of soft objects that are commonly seen in robotic surgeries, proving the robustness of the proposed method.

% This paper is structured as following: related work, method, experiment, limitation and conclusion.

% \subsection{Model-based Deformable Manipulation}

\begin{figure*}[t]
    \centering
    \includegraphics[width=0.95\linewidth]{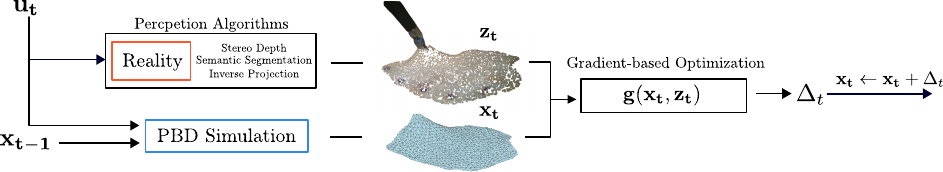} 
    \caption{A flow chart of the proposed residual mapping module in the simulation loop. At each time step, a control $\mathbf{u_t}$ is applied to both the real tissue and the simulation. In response, the PBD simulation solves for $\mathbf{x_t}$. A perception pipeline processes imagery data to obtain a surface point cloud $\mathbf{z_t}$ of the tissue. The residual mapping module estimates the residual deformation $\Delta_t$ via optimization, which is then used to update the simulation state.}
    % A perception pipeline will process imagery data to obtain a surface point cloud $\mathbf{z_t}$ of the tissue surface. A residual mapping module is used to identified the real-to-sim gap by predicting a residual deformation $\Delta_t$ (Section \ref{sec:mapping}). Then the detected gap is used to correct the simulation state's position and update the stiffness parameters. \textcolor{red}{TODO: change this to only reflect the residual mapping module
    \label{fig:residual_mapping_in_simulation_loop}
    \vspace{-1.5em}
\end{figure*}

\section{METHOD}

% \subsection{Preliminary: Position-based Dynamics Simulation}

% \begin{algorithm}[h]
% \centering
% \caption{PBD Algorithm}
% \label{pbd_algorithm}
% \begin{algorithmic}
% \State $x \gets 42$ \Comment{Placeholder, algorithm will be filled in later}
% \end{algorithmic}
% \end{algorithm}

% \begin{itemize}
%     \item Introduce history and property of PBD
%     \item The PBD simulation is given in Algorithm \ref{pbd_algorithm}. Given a few detail about the algorithm.
%     \item In this work, we focus on compensating the gap between PBD and real observation.
% \end{itemize}

\subsection{Problem Formulation}
Let the state of a soft body in a physics simulation at time stamp $t$ be $\mathbf{x_t} \in \mathbb{R}^{n \times 3}$, where $n$ is the number of particles of a mesh representing the soft body. We use the surface point cloud projected from the stereo-depth estimation as the observation, denoted as $\mathbf{z_t} \in \mathbb{R}^{m \times 3}$, where $m$ is the number of points in the point cloud observation. Let $\mathbf{u}_t$ be a point-based positional control that is applied to the real object and simulation simultaneously.
In this work, we use an extended position-based dynamic (PBD) simulator, which formulates constraints with positional and geometric data. A constraints-based formulation of PBD is given as 
% A control on the real object is carried out by grasping a point on the soft body and moving to a new location $\mathbf{p}_t$. In the simulation, $\mathbf{u}_t = (\text{idx}, \mathbf{p}_t)$ as a tuple of a pointer index and a location. It represents moving a single simulation state associated with the pointer to location $\mathbf{p}_t$. 
\begin{equation}\label{eqn:pbd_sim}
    \begin{split}
        \mathbf{x_t} = \mathcal{PBD}(\mathbf{x_{t-1}}&,\mathbf{u_{t}}, \mathbf{\Gamma},\mathbf{C}, \mathbf{k_c})\\
        s.t., \mathbf{C}(\mathbf{x_t}) &= \mathbf{0},\ \mathbf{\Gamma}(\mathbf{x_t}) = \mathbf{0}\\
        % \in \mathbb{R}
    \end{split}
\end{equation}
% where $\mathbf{x_0}$ is the initial configurations of the simulated mesh (e.g. resting length for a spring). 
where $\mathbf{\Gamma}$ are static (boundary) conditions that are enforced at each simulation step. The set of geometric constraints that define the deformation across the simulation is $\mathbf{C}(\mathbf{x}) = [C_1(\mathbf{x}),C_2(\mathbf{x}),\cdot\cdot\cdot, C_I(\mathbf{x})]^\top$. $\mathbf{k_c}\in\mathbb{R}^I$ is the set of weighting parameters associated with any kind of non-boundary constraint. One can interpret that $\{\mathbf{C, k_c}\}$ jointly defines the total energy potential (unitless) generated by constraints in a simulation:
% \begin{equation*}
%     \begin{split}
%         \mathbf{f}_{elastic} = -\nabla \mathbf{C}^T \mathbf{K_c}\mathbf{C}
%     \end{split}
% \end{equation*}
\begin{equation}\small
    \begin{split}
        \mathbf{U}(\mathbf{x}) = \large \sfrac{1}{2}\ \mathbf{C}(\mathbf{x})^{\top}\mathtt{diag}(\mathbf{k_c})\mathbf{C}(\mathbf{x})
    \end{split}
\end{equation}
This is used by PBD to iteratively update particle positions, by minimizing this energy term:
\begin{equation}\small
    \begin{split}
        \Delta \mathbf{x} &= \mathbf{M}^{-1} \nabla \mathbf{C}^\top \Delta \lambda\\
        \Delta \mathbf{\lambda} &= - \Bigl( \nabla \mathbf{C} \mathbf{M}^{-1} \nabla^{\top} \mathbf{C} + \mathbf{\tilde \alpha}\Bigr)^{-1} \Bigl(\mathbf{C} + \mathbf{\tilde \alpha} \mathbf{\lambda} \Bigr) \\
        \mathbf{\tilde \alpha} &= \mathtt{diag}(\mathbf{k_c})^{-1} / \Delta t^2)
    \end{split}
\end{equation}
in which $\mathbf{M}$ is a diagonal mass matrix and $\lambda$ is a Lagrange multiplier vector~\cite{macklin2016xpbd}. %In practice, the update is performed with the Gauss-Seidel method.}
Note, while we demonstrate the proposed method with PBD simulation, it can be extended to real-to-sim problems in other differentiable simulators (e.g., FEM, projective dynamics \cite{Du_2021_diffpd}) without losing generality. We are only including the simulation step for PBD so one can follow the use of parameters through a forward simulation.

In this work, we aim at optimizing stiffness parameters associated with each particle, $\mathbf{k} \in \mathbb{R}^n$. This converts to elastic constraints' weights in the simulation, denoted by $\mathbf{k_d}\subset \mathbf{k_c}$, by averaging stiffnesses across all involved particles. The value of $k_{d,i}\in \mathbf{k_{d}}$, that is, the weights of an elastic constraint $C_i(\mathbf{x}) \in \mathbf{C}$, is  
\begin{equation*}\small
    \begin{split}
        k_{d,i} = \Large \sfrac{1}{\mathtt{card}(Q_i)}\  \sum\nolimits_{q\in Q_i} k^q,\ k^q\subset \mathbf{k} 
    \end{split}
\end{equation*}
where particle indices $Q_i$ are considered by $C_i(\mathbf{x})$. Here, $Q_i$ contains $\mathtt{card}(Q_i)$ (cardinality) number of particles that are connected via mesh edges, triangles, and tetrahedrons.
This achieves non-homogeneous elastic stiffness across different regions of a soft body. Given that we generate a discretized mesh solely at the initial step, without engaging in any re-meshing procedures throughout the simulation, this imposes challenges on accurately representing isotropic material properties within the tissue. In light of this, we propose considering $\mathbf{k}$ as a spatially-variable stiffness.

% Our main interests in this work are 
% \begin{itemize}
%     \item identifying a residual mapping $\Delta_t(???)$ that quantifies the real-to-sim gap for deformable body, and apply the residual mapping to compensate the simulation states. 
%     \item optimizing the stiffness parameters online to reduce the real-to-sim gap.
% \end{itemize}
We apply the developed framework on two types of deformable bodies: \textit{thin-shell} objects that are simulated with a single layer of particles (similar to cloth) and \textit{volumetric} objects that are simulated with tetrahedral meshes. 

\subsection{Real-To-Sim Residual Mapping}\label{sec:mapping}
We will learn a residual mapping module that is incorporated into the simulation loop to estimate the real-to-sim gap. This module will enable matching between simulation particles to a point cloud observation as shown in Fig.~\ref{fig:residual_mapping_in_simulation_loop}. %The residual deformation can also be considered as a spatial real-to-sim gap. We use an optimization approach to quantify this gap. 
The method employs a gradient-based non-rigid point cloud registration to estimate the residual deformation $\Delta_t \in \mathbb{R}^{3\times n}$. To make the predicted residual mapping accurate and physically realistic, we consider both point cloud similarity $\mathcal{D}(\cdot)$ and physical realness $\mathcal{E}(\cdot)$ as cost functions to minimize. Both metrics are discussed in depth later.

Because the correspondence between the simulation particles and point cloud observation is unknown, we use Chamfer Distance as a measurement of similarities between two point clouds, which is computed by summing the squared distances between the nearest neighbor of two point clouds. For a thin-shell simulation, it is defined as 
\begin{equation}\small
    \begin{split}
        \mathcal{D}(\mathbf{x_t}, \mathbf{z_t}) &= \sum_{x\in \mathbf{x_t}} \min_{z\in \mathbf{z_t}}\|x - z\|^2_2  + \sum_{z\in \mathbf{z_t}} \min_{x\in \mathbf{x_t}}\|z -  x\|^2_2\\
    \end{split}\label{eqn:chamfer}
\end{equation}
A volumetric mesh can't be directly aligned to a surface point cloud observation by minimizing Equation \ref{eqn:chamfer} as internal particles are not observed. Therefore, we alternatively minimize $\mathcal{D}(\mathbf{x'_t}, \mathbf{z_t})$ where $\mathbf{x'_t} \in \mathbf{x_t}$ is the surface particles of the volumetric mesh, which is available from mesh initialization. 
\begin{figure}[!t]
    \centering
    \includegraphics[width=\linewidth]{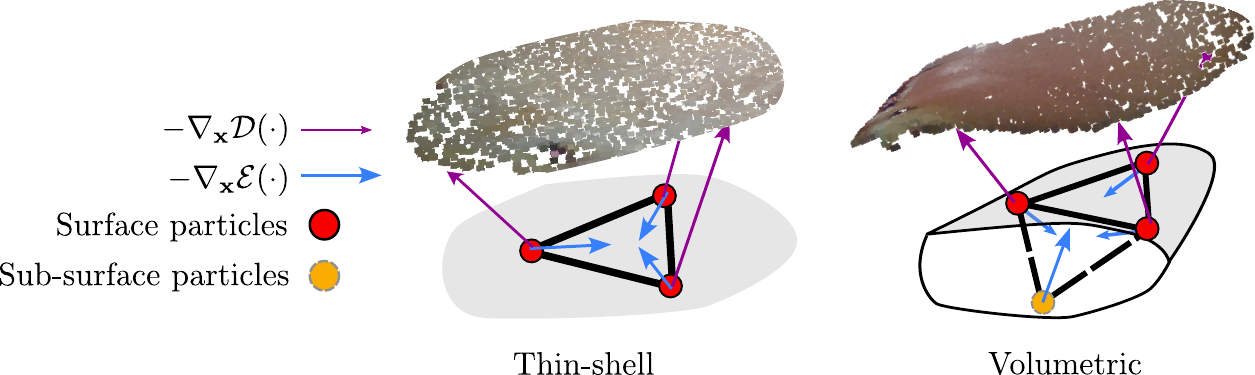} 
    \caption{Visualizations of the proposed residual mapping module for thin-shell and volumetric objects. Purple and blue arrows represent deformation due to optimizing $\mathcal{D}(\cdot)$ and $\mathcal{E}(\cdot)$, respectively. For sub-surface particles that are unobservable to the camera, only $\mathcal{E}(\cdot)$ informs how it will deform.}
    \label{fig:residual_optimization_vis}
    \vspace{-1.5em}
\end{figure}

We enforce physical realness $\mathcal{E}(\cdot)$ utilizing the simulator's geometric constraints. This is achieved by directly minimizing the simulation's energy potential:
\begin{equation}\small
    \begin{split}
       \mathcal{E}(\mathbf{x_t}) = \large \sfrac{1}{2}\ \mathbf{C}(\mathbf{x_t})^\top \mathtt{diag}(\mathbf{k_c'}) \mathbf{C}(\mathbf{x_t})
    \end{split}
\end{equation}
where $\mathbf{k_c'}$ is an uniform user-defined weight matrix and $\mathbf{C}$ are PBD constraints. Note that because sub-surface particles are not observable for volumetric meshes, this term is also essential for inferring their deformation. A visualization of how $\mathcal{D}(\cdot)$ and $\mathcal{E}(\cdot)$ affect the mesh deformation is shown in Fig.~\ref{fig:residual_optimization_vis}. Finally, a residual mapping function $\mathbf{g}(\mathbf{x_t}, \mathbf{z_t})$ is defined as
\begin{equation}
\footnotesize
    \begin{split}
    \mathbf{g}(\cdot)=
    \begin{cases}
        \text{argmin}_{\Delta_t} \mathcal{D}(\mathbf{x_t} + \Delta_t, \mathbf{z_t}) + \mathcal{E}(\mathbf{x_t}+\Delta_t) &\ \  \textit{if thin-shell,} \\
        \text{argmin}_{\Delta_t} \mathcal{D}(\mathbf{x'_t} + \Delta'_t, \mathbf{z_t}) + \mathcal{E}(\mathbf{x_t}+\Delta_t) &\ \ \textit{if volume.}\\
    \end{cases}
    \end{split}
\end{equation}
The above minimization problem is solved by performing 30 gradient descent steps with a learning rate of 50. At the end of a simulation step, $\Delta_t$ is used to update the current simulation state $\mathbf{x_t}$. Later, $\Delta_t$ is also viewed as a proxy to the real-to-sim gap that we seek to minimize through an online stiffness optimization approach.

\subsection{Online Stiffness Optimization}\label{sec:online_si}
The proposed online optimization method differs from previous real-to-sim methods as it does not rely on training on previously collected trajectories. In this way, we are solving an \textit{online} problem that is much more generalizable to real-world, unstructured environments. The algorithm is also embedded in the simulation loop as shown in Alg.~\ref{alg:online_si}.

One term we want to minimize directly is the residual gap, which characterizes how much the simulation deviates from the observation for the current time step. 
\begin{equation}
    \begin{split}
        \mathcal{L}_{gap} = \| \mathbf{g}(\mathbf{x_t}, \mathbf{z_t})\|
    \end{split}
\end{equation}
Its partial derivative with respect to stiffness parameters $\mathbf{k}$ is 
\begin{equation*}\small
\begin{split}
    \frac{\partial \mathcal{L}_{gap}}{\partial \mathbf{k}} 
    & = \frac{\partial \| \mathbf{g}(\mathbf{x_t}, \mathbf{z_t})\|}{\partial \mathbf{x_t}}\frac{\partial \mathcal{PBD}(\mathbf{x_{t-1}},..., \mathbf{k_c})}{\partial \mathbf{k_c}} \frac{\partial \mathbf{k_c}}{\partial \mathbf{k}}
\end{split}
\end{equation*}
% & = \frac{\partial \| \mathbf{g}(\mathbf{x_t}, \mathbf{z_t})\|}{\partial \mathbf{x_t}}\frac{\partial \mathbf{x_t}}{\partial \mathbf{k}} \\
where the first term is available by differentiating through the residual mapping module, and the second term is obtained by differentiating through a simulation step in Equation \ref{eqn:pbd_sim}. 

The residual gap alone doesn't consider historical information and, therefore, is sensitive to current observation noise. To address that, we introduce a history term computed over a set of previous time points $H_t$. It is defined as
\begin{equation}\small
    \begin{split}
        \mathcal{L}_{hist} = \sum\nolimits_{h\in H_t}\|\mathbf{x}_h + \Delta_h - \mathcal{PBD}(\mathbf{x}_h + \Delta_h, \mathbf{0}, \mathbf{C}, \mathbf{k_c})\|
    \end{split}
\end{equation}
where $\mathbf{x}_h$ and $\Delta_h$ are snapshots of simulation states and residual mapping at time point $h$. Specifically, $H_t$ is constructed by uniformly sampling four snapshots from a window of the closest 20 previous frames. This loss function finds the stiffness parameters that keep each snapshot at rest when no control is provided  (i.e., balancing external forces and internal elastic forces). This is a reasonable assumption when the system is not moving quickly.

\begin{algorithm}[!t]
\footnotesize
    \caption{Residual Mapping and Online Stiffness Optimization in a PBD simulation}
    \label{alg:online_si}
    \SetKwInOut{Input}{Input}
    \SetKwInOut{Output}{Output}

    \Input{Predefined control sequence $\mathbf{U}$, \text{stiffness} $\mathbf{k}$, real tissue $R$, \text{observation model} $\mathcal{H}(R)$, residual mapping module $\mathbf{g}.$}
        $\mathbf{z_0} \gets \mathcal{H}(R)$

        \tcp{Initialize a mesh and constraints.}
        $ \mathbf{x_0}, \mathbf{C} \gets \mathtt{initializeSimulation}(\mathbf{z_0})$ 
        
        $H_t \gets [\ ]$
        
        \For{\textbf{each} $\mathbf{u_t}\in \mathbf{U}$}{
            $ R \gets \textit{ApplyControl}(R, \mathbf{u_t})$
            
             $\displaystyle\mathbf{x_t} \gets \mathcal{PBD}(\mathbf{x_{t-1}, u_{t}, C, k_c})$
    
            $\mathbf{z_t} \gets \mathcal{H}(R)$
            
            $\Delta_t \gets \mathbf{g}(\mathbf{x_t}, \mathbf{z_t})$

            $\mathcal{L}_{gap} \gets \| \Delta_t\|$

            $ \mathcal{L}_{hist} \gets \sum_{h\in H_t}\|\mathbf{x}_h + \Delta_h - \mathcal{PBD}(\mathbf{x}_h + \Delta_h, \mathbf{0}, \mathbf{C}, \mathbf{k_c})\|$

            $\mathcal{L}_{smooth} \gets \frac{1}{2F}\sum_{f\in F} \sum_{i\in f} \sum_{j \in f}  k_i - k_j $

            $\mathcal{L}_{total} \gets \mathcal{L}_{gap} + \mathcal{L}_{hist} + \mathcal{L}_{smooth}$  
    % \State $\mathcal{L}_{gap} \gets \| \min_{\Delta_t} \mathcal{D}(\mathbf{x_t} + \Delta_t, \mathbf{z_t}) + \mathcal{E}(\mathbf{x_t}+\Delta_t) \|$
    
            $\displaystyle \mathbf{k} \gets \mathtt{Optimize}(\mathbf{k}, \nabla_\mathbf{k} \mathcal{L}_{total})$
    
            $H_t \gets [H_t, \mathbf{x_t} + \Delta_t]$
    
            $\mathbf{x_t} \gets \mathbf{x_t} + \Delta_t$
        }

        % \Return{$\displaystyle{{\partial \mathcal{L}}/{\partial \boldsymbol{\lambda}^{t}} }$}\\
\end{algorithm}

% \begin{algorithm}[t]
% \setstretch{1.1}
% \caption{Residual Mapping and Online Stiffness 
%  Optimization with a PBD simulation}\label{alg:online_si}
% \label{alg:online_si}
% \begin{algorithmic}[1]
% \Require Predefined control sequence $\mathbf{U}$, \text{stiffness} $\mathbf{k}$, real tissue $R$, \text{observation model} $\mathcal{H}(R)$, residual mapping module $\mathbf{g}.$
% \State $\mathbf{z_0} \gets \mathcal{H}(R)$
% \State $ \mathbf{x_0}, \mathbf{C} \gets \textit{InitializeSimulation}(\mathbf{z_0})$ 
% \State $H_t \gets [\ ]$
% % \While {$G(\mathbf{x}) < \varepsilon$}
% \For {\textbf{each} $\mathbf{u_t}\in \mathbf{U}$}
%     \State $ R \gets \textit{ApplyControl}(R, \mathbf{u_t})$
%     \State $\mathbf{x_t} \gets \mathcal{PBD}(\mathbf{x_{t-1}, u_{t}, C, K})$
%     \State $\mathbf{z_t} \gets \mathcal{H}(R)$
%     \State $\Delta_t \gets \mathbf{g}(\mathbf{x_t}, \mathbf{z_t})$
%     \State $\mathcal{L}_{total} \gets Computeloss(\Delta_t, H_t, \mathbf{k})$
%     % \State $\mathcal{L}_{gap} \gets \| \min_{\Delta_t} \mathcal{D}(\mathbf{x_t} + \Delta_t, \mathbf{z_t}) + \mathcal{E}(\mathbf{x_t}+\Delta_t) \|$
    
%     \State $\displaystyle \mathbf{k} \gets \textit{Optimize}(\mathbf{k}, \nabla_\mathbf{k} \mathcal{L}_{total})$
%     \State $H_t \gets [H_t, \mathbf{x_t} + \Delta_t]$
%     \State $\mathbf{x_t} \gets \mathbf{x_t} + \Delta_t$

% \EndFor
% \end{algorithmic}
% \end{algorithm}
In addition, we encourage a smooth spatial stiffness distribution by penalizing $\mathbf{k}$'s differences between neighboring particles. Let $F$ be a set of all faces or tetrahedrons that sorts tuples of particle indices, the smoothness loss is written as
\begin{equation}
    \begin{split}
       \mathcal{L}_{smooth} =  \frac{1}{2F}\sum_{f\in F} \sum_{i\in f} \sum_{j \in f}  k_i - k_j 
    \end{split}
\end{equation}
where $k_i \in \mathbf{k}$ is the stiffness value of the \textit{i}-th simulation particle. Note that intersection of different types of tissues is more complicated. which may be addressed by encouraging smoothness within individual semantic contours \cite{lin2023semanticsuper}. Lastly, all terms are summed up and back-propagated to the stiffness parameters. They are updated by taking a stochastic gradient descent step in every simulation step.

\section{EXPERIMENTS \& RESULTS}

% \subsubsection{Data Generation}
% The proposed real-to-sim mapping module is trained with both synthetic and real data. After initializing a soft body simulation and its boundary conditions, random surface vertices are picked as control points. A random simple linear control trajectory is generated for the control points to follow overtime. The simulation state $\mathbf{x_t}$ and simulated depth image re-projection $\mathbf{z_t}$ are recorded overtime. Similarly for real data, we apply several control sequences while simulating the control in the PBD simulation at the same time. Surface point cloud observations are acquired with a stereo camera system. The aforementioned unsupervised training strategy is then used to fine-tune the real-to-sim mapping module. During training, data of any two time points from the same control sequence is picked. We apply random noise and rotation as data augmentation to the synthetic data.
\begin{figure}[!t]
    \centering
    \includegraphics[width=\linewidth]{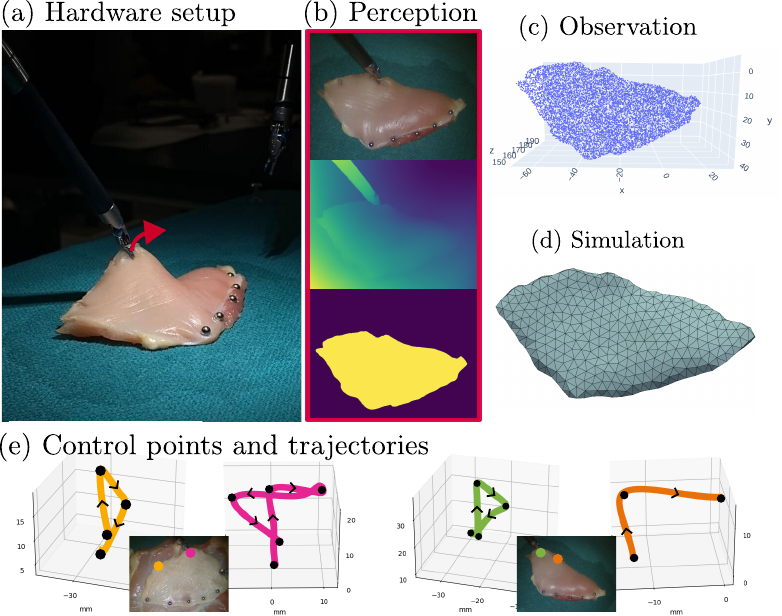} \vspace{-1.5em}
    \caption{The real-to-sim experiment setup in this work. (a): a piece of chicken muscle manipulated by a dVRK manipulator.
    (b): A perception pipeline estimates depth and a semantic mask of the tissue. (c): surface point cloud is generated by a camera inverse projection. (d): A simulation mesh is created with the initial observation. (e): four manipulation trajectories are visualized with their starting point labeled on real images.}
    \label{fig:experiment_setup}
    \vspace{-2em}
\end{figure}

\subsection{Real Experiment Setup}
 Fig.~\ref{fig:experiment_setup} shows our physical experiment setup. Our experimental procedures involve the utilization of the da Vinci Research Kit (dVRK) \cite{6907809}, employing its robotic gripper to precisely manipulate soft tissue by executing predefined trajectories. The trajectories are shown in Fig.~\ref{fig:experiment_setup} (e). Simultaneously, a stereo reconstruction pipeline processes stereo images captured by the da Vinci endoscopic camera in 720p, producing tissue surface point clouds. A piece of chicken skin and chicken muscle are used as subjects in our thin-shell and volumetric experiments, respectively. They are roughly 3mm and 2 cm thick. In both experiments, we used metal pins to fix one side of the tissue onto the bottom plane. 

We utilize the Raft-Stereo \cite{lipson2021raft} for stereo disparity estimation. Segment-Anything \cite{kirillov2023segany} aids in identifying image pixels corresponding to the tissue, allowing us to extract the tissue's surface point cloud from depth images. Subsequently, we employ inverse camera projection to convert the segmented depth into 3D positions. We also employ ArUco markers to determine the camera-to-world transformation. The point cloud is down-sampled to a size of 9000 points. Meshes are reconstructed by applying a ball-pivoting algorithm on the initial surface point cloud observation. For volumetric experiments, we process the surface meshes by adding thickness using the \textit{solidify} modifier in Blender. Meshes are re-meshed to contain 600 particles. The simulation's boundary conditions are selected at the locations of pins. The proposed residual mapping module requires a knowledge of surface particles that are visible from the camera. While we manually selected them, it is possible to select them automatically through visibility culling. The robotic gripper trajectories are manually labeled on the image.

We collect in total four trajectories, two for thin-shell and two for volumetric experiments. Later, they are referred as Thin-shell-1, Thin-shell-2, Volumetric-1, and Volumetric-2.

\begin{figure}[!t]
    \centering
    \includegraphics[width=\linewidth]{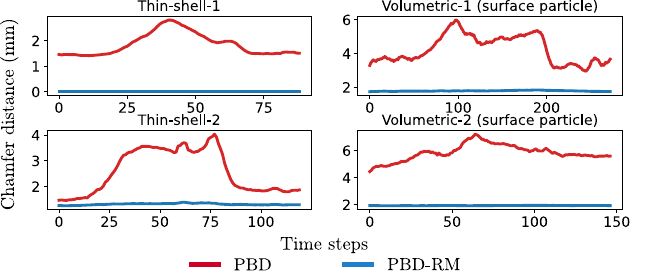}\vspace{-1em} 
    \caption{Comparison of the real-to-sim Chamfer distance (smoothed) between PBD and PBD-RM. In all four trajectories, the residual mapping module significantly reduces the Chamfer distance.}\label{fig:residual_result1}
    \vspace{1em}
    \includegraphics[width=\linewidth]{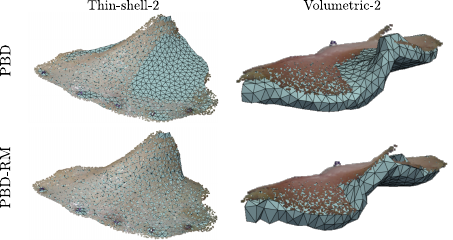}\vspace{-1.em}
    \caption{Surface point cloud observations overlaid on simulation meshes. With the residual mapping, the simulation mesh matches to textured point cloud observation better than the original PBD.}\label{fig:residual_result2}
    \vspace{-2em}
\end{figure}

\begin{figure}[!t]
    \centering
    \includegraphics[width=\linewidth]{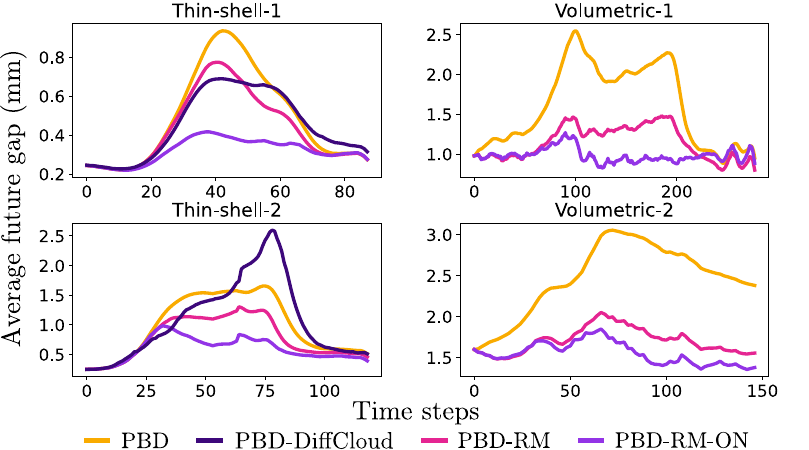} 
    \caption{Average future gap of comparison methods over time (smoothed). Initialized with $\mathbf{k}^2$, both our proposed components, residual mapping and online stiffness optimization, show large error reduction over PBD.}
    \label{fig:future_error1}
    \vspace{0.1cm}
    \includegraphics[width=\linewidth]{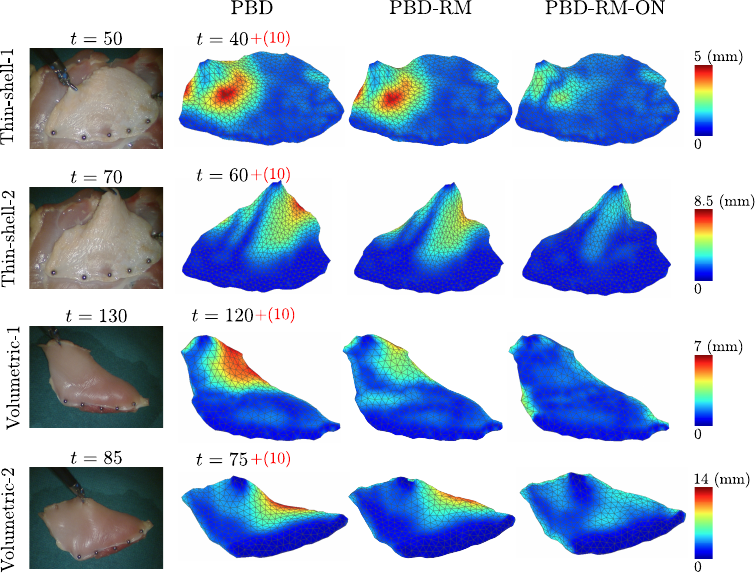} 
    \caption{Comparison of future gaps at selected time points. The gaps are visualized at $t=a+(b)$ time steps, meaning using the stiffness parameters at time $a$, forward the simulation for $b$ steps and then compute a gap. Both proposed PBD-RM and PBD-RM-ON can better predict the tissue's future deformation behavior than the original PBD simulation.}
    \label{fig:future_error2}
    \vspace{-1.5em}
\end{figure}
\subsection{Implementation Details}\label{sec:implementation_detail}
 This framework is implemented in Pytorch \cite{paszke2019pytorch}. Our PBD simulation adopts three types of geometric constraints: distance, volumetric, and shape-matching. They respectively preserve the distance between connected simulation particles, the volume of tetrahedral, and shapes formed by neighboring particles. We use distance and shape-matching constraints when simulating thin-shell tissue, whereas all three are used for volumetric tissue. For consistency, we optimize distance and shape-matching parameters for both thin-shell and volumetric test cases while keep volumetric constraint parameter fixed as ${\footnotesize k_{vol} = 1e10}$. In addition, we impose bounds on our simulation parameters with a \texttt{sigmoid} operation. These bounds are ${\footnotesize k_{dist}\in[0, 10]}$ and ${\footnotesize k_{shape}\in[0, 0.02]}$, which are empirically found effective to optimize in. As optimization-based methods are known to be sensitive to initialization, we evaluate the proposed method with three sets of initial parameters. They are ${\footnotesize \mathbf{k}^1:\{k_{dist}=5, k_{shape}=0.15\} }$, ${\footnotesize \mathbf{k}^2:\{k_{dist}=1, k_{shape}=0.1\}}$, ${\footnotesize \mathbf{k}^3:\{k_{dist}=0.2,k_{shape}=}$ ${\footnotesize {0.005}\}}$. All stiffness parameters are uniformly initialized. Note that initializations can be determine from prior knowledge of the deformable object. Here, we aim at testing every methods thoroughly.  They are optimized with an Adam optimizer with a learning rate of 0.1. Later, we refer to a PBD simulation with a residual mapping module as PBD-RM and one that further performs online updates as PBD-RM-ON. 

\begin{table*}[t!]
\setlength\tabcolsep{1.8em}
\centering
\caption{Comparison of time-average ($e_t$, $f_t$) of different methods with pre-defined initializations. Residual mapping ({PBD-RM}) and online optimization ({PBD-RM-ON}) improve prediction capabilities. }
\begin{adjustbox}{width=0.85\textwidth}
    \begin{tabular}{l|ccc|ccc}
    \toprule
    % \multirow{2}{*}{Velocity}
  \multirow{2}{*}{Methods} & $\mathbf{k}^1$ & $\mathbf{k}^2$  & $\mathbf{k}^3$ & $\mathbf{k}^1$ & $\mathbf{k}^2$ & $\mathbf{k}^3$ \\
 &\multicolumn{3}{c}{Thin-shell-1}  & \multicolumn{3}{c}{Thin-shell-2} \\ 

 \midrule
PBD & $0.52, 2.24$ & $0.47, 2.14$ & $0.34, 1.74$ & $1.06, 3.80$ &$0.96, 3.63$ &$0.89, 3.51$\\
PBD-DiffCloud& $0.45, \mathbf{1.76}$ & $0.46, 1.73$  & $0.33, \mathbf{1.57}$ & $0.90, 4.22$ & $0.97, 4.14$ & $1.02, 4.28$\\
PBD-RM & $0.46, 2.23$ & $0.42, 2.13$ & $0.32, 1.77$ & $0.84, 3.65$ & $0.76, 3.47$ & $0.67, 3.23$ \\
PBD-RM-ON & $\mathbf{0.40}, 2.04$  & $\mathbf{0.31}, \mathbf{1.66}$  & $\mathbf{0.30}, 1.72$ &  $\mathbf{0.65, 2.98}$ & $\mathbf{0.58, 2.72}$ & $\mathbf{0.62, 2.89}$ \\
% OFF-PBD-RM-ON & -  & -  & - & - & - & -\\
    \midrule
 &\multicolumn{3}{c}{Volumetric-1}  & \multicolumn{3}{c}{Volumetric-2} \\ \midrule
PBD & $1.20, 3.15$ & $1.59, 4.03$ & $2.51, 6.57$ & $1.75, 4.45$ & $2.46, 5.53$ & $3.33, 7.58$\\
PBD-RM & $1.02, 3.06$ & $1.13, 3.52$ & $1.30, 4.32$ & $1.41, 4.33$ & $1.65, 5.00$ & $1.86, 5.70$\\
PBD-RM-ON &  $\mathbf{0.94, 2.91}$ & $\mathbf{0.97, 3.21}$  & $\mathbf{1.00, 3.38}$ & $\mathbf{1.40, 4.22}$ & $\mathbf{1.48, 4.56}$ & $\mathbf{1.52, 4.71}$\\
\bottomrule
\end{tabular}
\end{adjustbox}
\label{tbl:main_result}
% \vspace{-em}
\end{table*}

\begin{figure*}[!t]
    \centering
    \includegraphics[width=\linewidth]{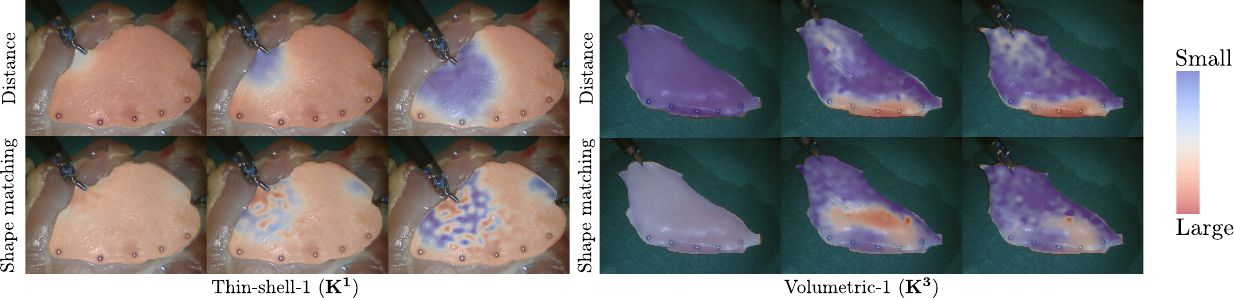}\vspace{-0.5em} 
    \caption{Projection of spatial distribution of optimized distance and shaping matching stiffness parameters over time. Red indicates a larger stiffness and blue indicates a smaller stiffness value. From left to right, stiffness distributions evolve over time.}
    \label{fig:stiffness_overlay}
    \vspace{-1.7em}
\end{figure*}

\subsection{Residual Mapping Evaluation}
In this section, we evaluate the effectiveness of applying the proposed residual mapping. The simulations are initialized with stiffness $\mathbf{k_C}^2$. We measure the error between simulation and observation with Chamfer distances. Fig.~\ref{fig:residual_result1} shows that the errors are notably reduced in all trajectories using the proposed residual mapping module. They are stably kept at low levels throughout all trajectories. In the thin-shell cases, the proposed module shrinks the average errors from 1.84 mm and 2.55 mm to 0.001 mm and 1.31 mm. Likewise, the average errors are reduced from 4.24 mm and 5.79 mm to 1.78 mm and 1.93 mm in the volumetric case. Fig.~\ref{fig:residual_result2} visualizes the reduced differences between point clouds and mesh with our mapping module. For the volumetric case, the module simultaneously aligns the volumetric mesh's surface particles to an observation and also deforms sub-surface particles, respecting PBD's geometric constraints.

\subsection{Online Stiffness Optimization Evaluation}
Here, we evaluate the method's future deformation prediction capabilities. Specifically, we formulate an average future gap $e_t$ and a future keypoint error $f_t$ as our metrics. Knowing the current simulation state $\mathbf{x_t}$ and a future control sequence, we can roll out a simulation, \textit{without knowing future observations}, to get future state sequence $[\mathbf{x_{t+1}}...\mathbf{x_{t+T}}]$. The average future gap is computed as
\begin{equation}\small
    \begin{split}
        e_{t} &= \large \sfrac{1}{T}\  \sum\nolimits_{s=1}^T \| \mathbf{g}(\mathbf{x_{t+s}}, \mathbf{z}_{t+s}) \|. \\
    \end{split}
\end{equation}
Let $\mathbf{p}_t$ be a vector of keypoints' positions at time $t$, and let $x^i_t$ represents the $i$-th particle in $\mathbf{x_t}$. The future keypoint error is defined as
\begin{equation}\small
    \begin{split}
        f_{t} &= \sum_{p_0, p_{t+T} \in \mathbf{p_0}, \mathbf{p}_{t+T}}\| p_{t+T} -  (\Delta p + p_0) \|\\
        \mathtt{where}\ \Delta p & \approx x^{nn}_{t+T} - x^{nn}_0\\
        nn &= \mathtt{NearestNeighIndex}(\mathbf{x_0}, p_0) \\ 
    \end{split}
\end{equation}
Here, keypoint displacement $\Delta p$ is approximated by the displacement of $p$'s closest neighboring particles at $t=0$. For all experiments, we pick a future horizon $T=10$. A set of 15 keypoints are manually labeled on images every 10 frames for each trajectory. These keypoints are in deformed regions and are selected based on their visual features.

For thin-shell experiments, we compare to DiffCloud \cite{sundaresan2022diffcloud}, a baseline that optimizes simulation parameters by minimizing point cloud differences on a training trajectory. we optimized with the first 30 frames of each trajectory for 25 epochs. Table \ref{tbl:main_result} shows a comparison on the two aforementioned metrics. Compared to the original PBD, PBD-RM shows reduced errors in all experiments, indicating that correcting the current simulation state with residual deformation leads to better future prediction. Compared to DiffCloud, PBD-RM-ON doesn't consistently outperform it on the relatively shorter and simpler Thin-shell-1. However, it performs significantly better than all other methods in the other three experiments. DiffCloud produces unsatisfactory results on Thin-shell-2 as it overfits to the beginning of that trajectory, whereas the proposed method does not as it updates parameters online. Fig.~\ref{fig:future_error1} supports the same observation, visualizing the average future gaps over time. Our PBD-RM-ON largely reduces the gap in the middle of trajectories where deformations are large. Fig.~\ref{fig:future_error2} visualizes the spatial distribution of future gaps. It indicates larger errors located where tissues are experiencing bending. Both PBD-RM and PBD-RM-ON are effective at reducing errors in those regions. In Fig.~\ref{fig:stiffness_overlay}, the spatial distribution of optimized parameters is overlaid on real images. The stiffness parameters exhibit spatial and temporal variations. Effects such as distance stiffness lowering around the deforming areas in the thin-shell case can be attributed to the chicken skin's inherent softness. In cases of heterogeneous parameter distributions, we hypothesize that their spatial variation may better approximate the complex behavior of the real tissue. However, whether the spatial variations represent can real material stiffness needs further future studies.

% \subsection{Thin-shell Manipulation}
\section{DISCUSSION \& CONCLUSION}
In this work, we have introduced an online simulation optimization framework designed to address the real-to-sim gap that arises in deformable tissue manipulation. A residual mapping module is seamlessly integrated into a simulation loop, achieving a minimal Chamfer distance between simulated particles and observation while preserving the geometric relationships in the simulator. Our optimization approach updates constraints' stiffness parameters online. In real tissue experiments, it is proven to be effective at improving predictive performance. This improvement holds the potential for enhancing model-based control for DOM, such as identifying an accurate simulation model for model predictive control in autonomous surgeries. For deploying this work into real applications, one limitation is its computation speed. Currently, PBD-RM-ON takes 0.9s and 2.6s to complete a step for thin-shell and volumetric meshes. Additional, our perception pipeline takes 0.4s per image. Despite that, it can be accelerated by GPU implementation or asynchronous computation. Other key issues, such as perception failure due to noise and unmodeled phyiscal effects such as other external foces, need to addressed in future works. A possible future avenue of this work would be learning residual non-linear constraints to capture more intricate tissue behaviors.

\section*{ACKNOWLEDGMENT}
This work is funded by the US Army Telemedicine and Advanced Technologies Research Center (TATRC) and NSF CAREER award \#2045803.
%%%%%%%%%%%%%%%%%%%%%%%%%%%%%%%%%%%%%%%%%%%%%%%%%%%%%%%%%%%%%%%%%%%%%%%%%%%%%%%%

\label{Bibliography}
%\lhead{\emph{Bibliography}} % Change the page header to say "Bibliography"
\bibliographystyle{unsrt} % unsrt % Use the "unsrtnat" BibTeX style for formatting the Bibliography
\bibliography{reference}

\end{document}